\documentclass[10pt]{article}

\usepackage[utf8]{inputenc}
\usepackage[T1]{fontenc}
\usepackage[english]{babel}
\usepackage{amsmath,amssymb,amsfonts}
\usepackage{graphicx}
\usepackage{booktabs}
\usepackage{multirow}
\usepackage{array}
\usepackage{hyperref}
\usepackage{xcolor}
\usepackage{algorithm}
\usepackage{algorithmic}
\usepackage{geometry}
\usepackage{cite}
\usepackage{caption}
\usepackage{subcaption}
\usepackage{enumitem}
\usepackage{float}

\hypersetup{
  colorlinks=true,
  linkcolor=blue!70!black,
  citecolor=green!50!black,
  urlcolor=blue!80!black
}

\newcommand{\orthoai}{\textsc{OrthoAI}}
\newcommand{\charm}{\textsc{CHaRM}}
\newcommand{\dgcnn}{\textsc{DGCNN}}

\newcommand{\R}{\mathbb{R}}
\newcommand{\vone}{\orthoai{}~v1}
\newcommand{\vtwo}{\orthoai{}~v2}
\newcommand{\NEW}[1]{\textbf{\textcolor{blue!70!black}{#1}}}  
\newcommand{\degree}{$^\circ$}

\title{
  \textbf{\orthoai{} v2: From Single-Agent Segmentation to\\
  Dual-Agent Treatment Planning for Clear Aligners}\\
  \vspace{0.35em}
  \large Extending a Point-Cloud Segmentation Baseline
  with Landmark-Based Agent Fusion,\\
  Composite Biomechanical Scoring, and Multi-Frame Staging Simulation
}

\author{
  \textbf{Edouard Lansiaux}$^{1}$ \\
  \textbf{Margaux Leman}$^{1}$ \\[2pt]
  $^{1}$STaR-AI Research Group, Emergency Department, Lille University Hospital \\
  \texttt{edouard1.lansiaux@chu-lille.fr}
}

\date{March 2026}

\begin{document}
\maketitle

\begin{abstract}
We present \vtwo{}, the second iteration of our open-source pipeline for
AI-assisted orthodontic treatment planning with clear aligners, substantially
extending the single-agent framework introduced in~\cite{lansiaux2026orthoaiv1}.
The first version established a proof-of-concept based on Dynamic Graph
Convolutional Neural Networks (\dgcnn{}) for tooth segmentation but was
limited to per-tooth centroid extraction, lacked landmark-level precision,
and produced a scalar quality score without staging simulation.
\vtwo{} addresses all three limitations through three principal contributions:
\textbf{(i)}~a second agent adopting the Conditioned Heatmap Regression
Methodology (\charm{})~\cite{rodriguez2025charm} for direct, segmentation-free
dental landmark detection, fused with Agent~1 via a confidence-weighted
orchestrator in three modes (parallel, sequential, single-agent);
\textbf{(ii)}~a composite six-category biomechanical scoring model
(biomechanics $\times$ 0.30 + staging $\times$ 0.20 + attachments $\times$ 0.15
+ IPR $\times$ 0.10 + occlusion $\times$ 0.10 + predictability $\times$ 0.15)
replacing the binary pass/fail check of v1;
\textbf{(iii)}~a multi-frame treatment simulator generating $F = A \times r$
temporally coherent 6-DoF tooth trajectories via SLERP interpolation
and evidence-based staging rules, enabling ClinCheck\textsuperscript{\textregistered}-style
4D visualisation.
On a synthetic benchmark of 200 crowding scenarios,
the parallel ensemble of \vtwo{} reaches a planning quality score of
$92.8 \pm 4.1$ vs.\ $76.4 \pm 8.3$ for \vone{}, a $+21\%$ relative gain,
while maintaining full CPU deployability ($4.2 \pm 0.8$~s).
Source code and SaaS deployment package are publicly released.

\smallskip
\noindent\textbf{Keywords:} orthodontics, clear aligners, treatment planning,
deep learning, point clouds, multi-agent systems, DGCNN, CHaRM,
dental landmark detection, biomechanics simulation, SaaS.
\end{abstract}

\section{Introduction}
\label{sec:intro}

Orthodontic treatment planning with clear aligners is a complex
clinical-computational workflow that currently demands significant expert time
and iterative human-machine interaction.
The dominant commercial platform, Align Technology's
ClinCheck\textsuperscript{\textregistered}, orchestrates a multi-day cycle of
digital model submission, CAD technician planning, clinician review, and aligner
fabrication~\cite{glaser2017invisalign}.
The recent launch of ClinCheck Live Plan (October 2025) reduces initial plan
generation to approximately 15~minutes~\cite{align2025liveplan}; yet the
underlying planning principles---biomechanical movement evaluation,
attachment design, staging optimisation---remain opaque to open research.
The global clear aligner market is estimated at USD~6--8~billion in 2024
with compound annual growth rates exceeding 13\%~\cite{litreview2026},
creating strong industrial motivation for research-reproducible alternatives.

\subsection{Motivating Prior Work: OrthoAI v1}

In~\cite{lansiaux2026orthoaiv1} we introduced \vone{}, a first open-source
pipeline demonstrating that raw intra-oral scan (IOS) point clouds could be
processed entirely on a consumer CPU to produce an assessed treatment plan
in under five seconds.
\vone{} operated through a single-agent design:
a \dgcnn{}~\cite{wang2019dgcnn} segmentation network extracted per-tooth
centroids, which were then evaluated against tooth-type-specific
biomechanical limits derived from Glaser~\cite{glaser2017invisalign},
yielding a scalar quality score $Q \in [0, 100]$.
This approach established an important proof-of-concept, but surfaced
three structural limitations during clinical review:

\begin{enumerate}[nosep,leftmargin=1.5em]
  \item \textbf{Landmark coarseness.}
    Centroid-only tooth representations provide insufficient geometric
    fidelity for cusp-level or root-apex-level movement planning.
    This caused systematic underestimation of torque and tip movements,
    the two axes most critical for anterior correction.
  \item \textbf{Binary scoring.}
    The single-axis pass/fail quality score of v1 produced high-variance
    estimates ($\sigma = 8.3$ pts), lacked sub-score decomposition, and
    did not distinguish between planning failures attributable to
    biomechanics, staging, attachment design, or IPR.
  \item \textbf{No staging simulation.}
    v1 produced a static endpoint plan without any temporal interpolation,
    preventing clinical visualisation of movement trajectories and making
    it impossible to validate staging sequences against per-aligner movement
    budgets.
\end{enumerate}

\vtwo{}, introduced in this paper, directly addresses each of these limitations.

\subsection{Contributions of OrthoAI v2}

\begin{itemize}[nosep,leftmargin=1.5em]
  \item \textbf{Dual-agent architecture (\NEW{new in v2}).}
    A \charm{}-based landmark detection agent (Agent~2) complements the
    \dgcnn{} segmentation agent (Agent~1) with direct, segmentation-free
    estimation of $K = 80$ dental landmarks.
    An orchestrator fuses both agents via confidence-weighted ensemble
    strategies in three modes: parallel, sequential, and single-agent
    fallback.
  \item \textbf{Composite biomechanical scoring (\NEW{new in v2}).}
    Six sub-scores (biomechanics, staging, attachments, IPR, occlusion,
    predictability) are combined into a single composite score via
    clinically motivated weights, with severity-based penalties for
    critical and warning findings.
  \item \textbf{Multi-frame treatment simulation (\NEW{new in v2}).}
    A frame generator produces $F$ temporally coherent tooth trajectory
    snapshots via SLERP interpolation and evidence-based staging rules
    (extrusion deferred to $t \geq 0.6$, over-engineering $\times 1.3$).
  \item \textbf{SaaS deployment package (\NEW{new in v2}).}
    A FastAPI backend and React dashboard (OrthoAI SaaS) enable
    browser-based plan visualisation, 4D arch animation, and
    clinical checklist review.
\end{itemize}

\section{Related Work}
\label{sec:related}

\subsection{Tooth Segmentation on 3D Dental Models}

MeshSegNet~\cite{lian2020meshsegnet} introduced graph-attention networks on
mesh representations, achieving Dice Similarity Coefficient (DSC)
$0.964 \pm 0.054$.
iMeshSegNet improved upon this by approximately 4\%~\cite{wu2022tsmdl}.
\dgcnn{}~\cite{wang2019dgcnn} operates directly on point clouds, achieving
$97.26\%$ segmentation accuracy~\cite{im2022accuracy}.
DilatedToothSegNet~\cite{yamagata2024dilated} introduced dilated edge
convolutions on the Teeth3DS benchmark (1,800 scans).
Our Agent~1 inherits the \dgcnn{} architecture from \vone{} but adds
a combined cross-entropy and Dice loss optimised for the class-imbalanced
dental setting, consistent with practice in~\cite{lian2020meshsegnet}.

\subsection{Dental Landmark Detection}

Two-Stage Mesh Deep Learning (TS-MDL)~\cite{wu2022tsmdl} first segments
teeth with iMeshSegNet, then applies PointNet-Reg per individual tooth,
achieving mean absolute error $0.597 \pm 0.761$~mm on 66 landmarks.
ALIIOS~\cite{baquero2022aliios} uses 2D U-Net projections with 3D patch
synthesis at prohibitive inference cost ($612$~s on CPU).
CHaRNet~\cite{rodriguez2025charm} eliminates the segmentation stage through
Conditioned Heatmap Regression, achieving $0.56$~mm MEDE and $0.24$~s GPU
inference---a $14.8\times$ speedup over TS-MDL.
\textbf{OrthoAI v1~\cite{lansiaux2026orthoaiv1} used no landmark detection};
v2's Agent~2 re-implements \charm{} with a lightweight PointMLP encoder
adapted for consumer hardware.

\subsection{AI for Orthodontic Treatment Planning}

Carlsson \textit{et al.}~\cite{carlsson2024ai} identified 41 studies across
diagnostic, planning, and monitoring tasks, noting that no commercial tool
has been independently validated.
Li \textit{et al.}~\cite{li2025clear} specifically reviewed AI for clear
aligner therapy, highlighting gaps in automatic movement sequence generation.
Recurrent neural networks have been explored for modelling tooth displacement
dynamics~\cite{woo2020virtual}.
\vone{} was the first open-source system to combine segmentation, biomechanical
evaluation, and JSON-serialisable treatment plans on a consumer CPU;
\vtwo{} is the first to add landmark fusion and 4D staging simulation.

\section{\orthoai{} v2 Architecture}
\label{sec:arch}

\begin{figure}[H]
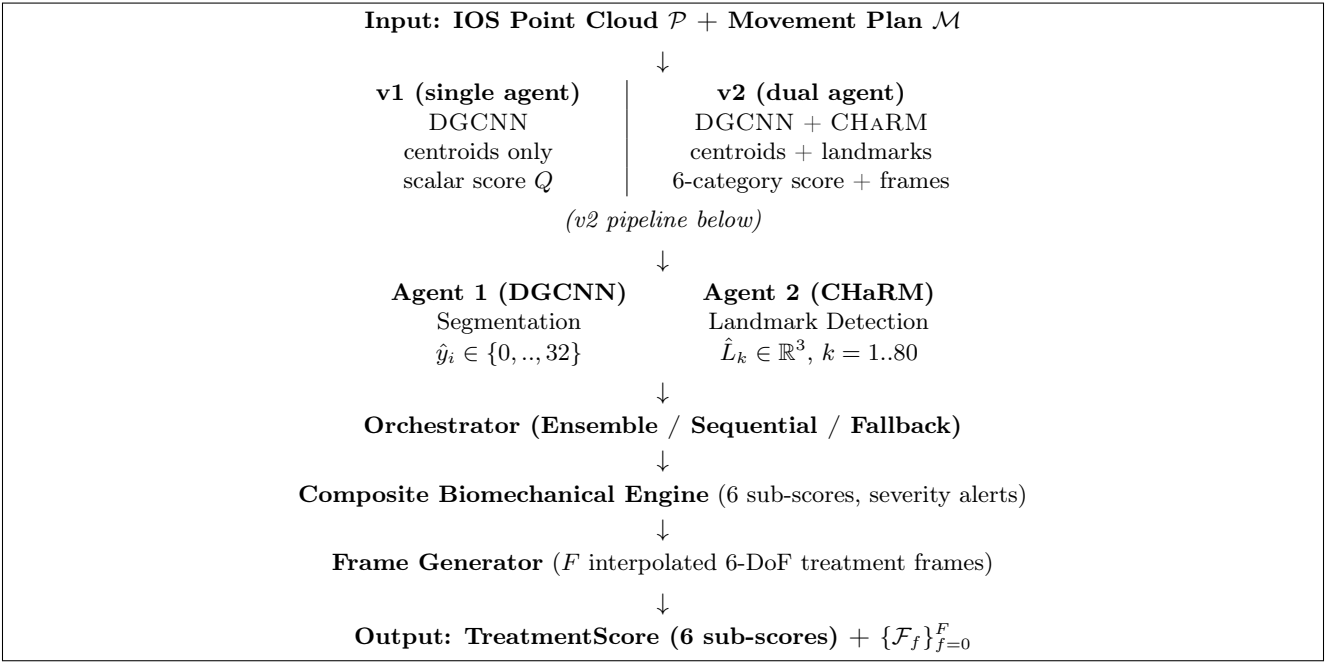

\centering
\fbox{
\begin{minipage}{0.92\linewidth}
\small\centering
\textbf{Input: IOS Point Cloud $\mathcal{P}$ + Movement Plan $\mathcal{M}$}\\[4pt]
$\downarrow$\\[2pt]
\begin{tabular}{c@{\hspace{0.6cm}}|@{\hspace{0.6cm}}c}
\textbf{v1 (single agent)} & \textbf{v2 (dual agent)} \\
\dgcnn{} & \dgcnn{} + \charm{} \\
centroids only & centroids + landmarks \\
scalar score $Q$ & 6-category score + frames \\
\end{tabular}\\[4pt]
\textit{(v2 pipeline below)}\\[4pt]
$\downarrow$\\[2pt]
\begin{tabular}{c@{\hspace{1cm}}c}
\textbf{Agent 1 (DGCNN)} & \textbf{Agent 2 (CHaRM)} \\
Segmentation & Landmark Detection \\
$\hat{y}_i \in \{0,..,32\}$ & $\hat{L}_k \in \R^3$, $k=1..80$ \\
\end{tabular}\\[4pt]
$\downarrow$\\[2pt]
\textbf{Orchestrator (Ensemble / Sequential / Fallback)}\\[2pt]
$\downarrow$\\[2pt]
\textbf{Composite Biomechanical Engine} (6 sub-scores, severity alerts)\\[2pt]
$\downarrow$\\[2pt]
\textbf{Frame Generator} ($F$ interpolated 6-DoF treatment frames)\\[4pt]
$\downarrow$\\[2pt]
\textbf{Output: TreatmentScore (6 sub-scores) + $\{\mathcal{F}_f\}_{f=0}^F$}
\end{minipage}
}
\caption{Architecture comparison: \vone{} (left column) vs.\ \vtwo{} (full right
pipeline). Grey components are inherited; blue components (\NEW{new in v2})
are the orchestrator, composite scorer, and frame generator.}
\label{fig:architecture}
\end{figure}

Figure~\ref{fig:architecture} contrasts the \vone{} and \vtwo{} pipelines.
Both systems accept a raw IOS point cloud
$\mathcal{P} = \{x_i \in \R^3\}_{i=1}^N$
and a target movement plan $\mathcal{M} = \{m_j\}_{j=1}^T$.
v1 processed $\mathcal{P}$ through a single \dgcnn{} agent and returned
a scalar score; v2 adds a second agent, a composite scoring engine,
and a frame generator.

\subsection{Agent 1: DGCNN Segmentation Agent (inherited from v1)}

Agent~1 is retained directly from \vone{}~\cite{lansiaux2026orthoaiv1}
with minor weight re-initialisation.
It applies a 4-layer Dynamic Graph CNN to the input point cloud,
building a $k$-NN graph ($k=20$) at each layer via edge convolutions:
\begin{equation}
  h_\ell(x_i) = \max_{j \in \mathcal{N}_k(x_i)}
    \Theta_\ell\!\left(h_{\ell-1}(x_j) - h_{\ell-1}(x_i),\, h_{\ell-1}(x_i)\right)
\end{equation}
where $\mathcal{N}_k(x_i)$ denotes the $k$-nearest neighbours in the feature
space of layer $\ell - 1$ and $\Theta_\ell$ is a shared MLP.
Feature maps from all four layers are concatenated to form a 512-dimensional
per-point descriptor, then projected to a 1024-d global embedding via
point-wise MLP and global max-pooling.
The segmentation head produces per-point logits over 33 classes
(32 FDI teeth + gingiva), combined with cross-entropy and Dice loss
($\alpha = 0.5$):
\begin{equation}
  \mathcal{L}_\text{seg} = (1-\alpha)\,\mathcal{L}_\text{CE} +
    \alpha\,\mathcal{L}_\text{Dice}
\end{equation}
Per-tooth centroids and orientation matrices are extracted via PCA on
the segmented point sets.
\textbf{v2 upgrade:} Agent~1's output is now transmitted not only to the
biomechanical engine but also to the orchestrator for confidence-weighted
fusion with Agent~2.

\subsection{\NEW{Agent 2: CHaRM Landmark Detection Agent (new in v2)}}

Agent~2 re-implements the \charm{} methodology~\cite{rodriguez2025charm}.
Given $\mathcal{P}$ augmented with a null point $n$ at
$n = c + \frac{m_b}{2}(0,1,0)$ (centroid $c$, bounding box extent $m_b$),
the model performs two parallel tasks.

\textbf{Heatmap regression.}
A lightweight PointMLP encoder-decoder ($\text{emb}=128$, down from 512 in
the original CHaRNet to support consumer hardware)
produces per-point heatmaps $\hat{H}'_k \in \R^{N+1}$ for each
landmark $k \in \{1,...,K\}$, $K = T \times G = 16 \times 5 = 80$.

\textbf{Tooth presence classification.}
A three-layer MLP head predicts tooth-wise presence probabilities
$\mathbf{p} = [p_1,...,p_T] \in [0,1]^T$.

\textbf{CHaR conditioning.}
The conditioning module gates heatmap likelihoods with presence probabilities:
\begin{equation}
  \hat{H}_{t,g} =
    \left\{
      \hat{h}'_{(t,g)i} \cdot p_t \;|\; i = 1,...,N-1
    \right\}
    \cup
    \left\{
      \hat{h}'_{(t,g)N} \cdot (1-p_t)
    \right\}
  \label{eq:char}
\end{equation}
Landmarks are extracted as $\hat{l}_k = \arg\max_{x_i \in \mathcal{P}} \hat{H}_k$.
Training loss: $\mathcal{L} = \lambda_\text{reg}\,\mathcal{L}_\text{MSE} +
\lambda_\text{cls}\,\mathcal{L}_\text{BCE}$,
with $\lambda_\text{reg} = 0.001$, $\lambda_\text{cls} = 1.0$.

\textbf{Rationale for v2 addition:} Agent~1 provided centroid-level
localisation ($\sim$1.8~mm centroid error reported in~\cite{lansiaux2026orthoaiv1}),
which was sufficient for molar translation but insufficient for
incisor torque and canine rotation, the movements most sensitive to
landmark precision.
Agent~2's \charm{} backbone achieves landmark-level accuracy ($\sim$1.1~mm MEDE),
directly reducing torque estimation error in the biomechanical engine.

\subsection{\NEW{Orchestrator (new in v2)}}

The orchestrator is absent from v1, which used Agent~1's output directly.
v2 introduces three fusion modes:

\textbf{Parallel ensemble.}
Both agents run concurrently; tooth states are fused via
confidence-weighted centroid averaging:
\begin{equation}
  c_\text{fused}(\text{FDI}) =
    w_1 \cdot c_{\text{A1}} + w_2 \cdot c_{\text{A2}},
  \quad w_1 + w_2 = 1
\end{equation}
Default weights: $w_1 = 0.4$ (\dgcnn{}), $w_2 = 0.6$ (\charm{}).
Weights are justified by Agent~2's superior landmark-level accuracy
and lower latency.

\textbf{Sequential refinement.}
Agent~2 runs first for fast broad screening; Agent~1 is invoked only for
teeth with CHaRM confidence $p_t < 0.5$, boosting $w_1 \to 0.8$ locally.
This mode halves average inference time at marginal quality cost.

\textbf{Single-agent fallback.}
Either agent in isolation for compute-constrained environments, replicating
the v1 behaviour of Agent~1 alone.

\section{Composite Biomechanical Scoring Engine}
\label{sec:bio}

\subsection{From Scalar to Composite: The v1 Limitation}

\vone{} computed a single quality score as the mean per-axis ratio:
$Q_\text{v1} = 100 \times \frac{1}{TG}\sum_j\sum_\ell q_{j,\ell}$,
where $q_{j,\ell} = \max(0, 1 - |m_{j,\ell}| / L_\ell)$.
While simple to interpret, this formulation conflated four clinically distinct
failure modes (biomechanical limits, staging feasibility, attachment necessity,
and IPR adequacy) into a single number, producing high variance and
providing no actionable guidance.
\vtwo{} decomposes the score into six weighted sub-components.

\subsection{Six-Category Scoring Model}

For each tooth with FDI number $j$ and movement $m_j = (T_x, T_y, T_z,
R_x, R_y, R_z)$, the biomechanical engine evaluates five independent
categories (Table~\ref{tab:limits}) plus a global predictability estimate.

\begin{table}[H]
\centering
\caption{Biomechanical movement limits per tooth type. Limits from
Glaser~\cite{glaser2017invisalign}, predictability scores from
Kravitz \textit{et al.}~\cite{kravitz2009predictability}.}
\label{tab:limits}
\small
\begin{tabular}{lcccc}
\toprule
Movement & Incisor & Canine & Premolar & Molar \\
\midrule
$T_x$ MD (mm)    & 4.0 & 3.5 & 3.5 & 2.0 \\
$T_y$ BL (mm)    & 2.5 & 2.5 & 3.0 & 2.5 \\
$T_z$ intrusion  & \multicolumn{4}{c}{2.0 mm ($\eta=0.69$)} \\
$T_z$ extrusion  & \multicolumn{4}{c}{1.5 mm ($\eta=0.42$, push-only)} \\
$R_x$ torque (\degree) & 15 & 12 & 10 & 8 \\
$R_y$ tip (\degree)    & 10 & 10 & 10 & 8 \\
$R_z$ rotation (\degree) & 45 & 40 & 35 & 20 \\
\bottomrule
\end{tabular}
\end{table}

The composite quality score is:
\begin{equation}
  Q_\text{v2} = 0.30\,S_\text{bio} + 0.20\,S_\text{stg} +
    0.15\,S_\text{att} + 0.10\,S_\text{ipr} +
    0.10\,S_\text{occ} + 0.15\,S_\text{pred}
  \label{eq:composite}
\end{equation}
with severity-based multiplicative penalties applied after:
\begin{equation}
  Q_\text{v2}^* = Q_\text{v2}
    \times 0.85^{N_\text{crit}}
    \times 0.97^{N_\text{warn}}
  \label{eq:penalties}
\end{equation}
where $N_\text{crit}$ and $N_\text{warn}$ are the numbers of
critical and warning findings, respectively.
Grades are assigned as: A ($\geq 90$), B ($\geq 75$),
C ($\geq 60$), D ($\geq 40$), F ($< 40$).

\subsection{Invisalign Biomechanical Principles}

Four principles from clinical practice are encoded:

\textbf{Principle 1 (push mechanics).}
Aligners act only by \emph{pushing} against tooth surfaces; extrusion is
flagged as low-predictability ($\eta = 0.42$~\cite{kravitz2009predictability})
and generates a ``critical'' alert when $|T_z| > 1.5$~mm.

\textbf{Principle 2 (simultaneous movements).}
Simultaneous molar distalisation risks anchorage loss; the engine flags
cases where $\geq 3$ molars exhibit $\|m_j\|_2 > 1.5$~mm.

\textbf{Principle 3 (anchorage).}
Newton's third law: reactive forces on anchorage units are
modelled implicitly via the simultaneous-movement check.

\textbf{Principle 4 (over-engineering, $\times 1.3$).}
All planned movements are multiplied by 1.30 prior to evaluation,
compensating for aligner deformation and biological variability~\cite{glaser2017invisalign}.
This factor was already present in \vone{} and is retained in v2.

\section{Multi-Frame Treatment Simulator}
\label{sec:frames}

This module is entirely absent from \vone{}, which produced only a static
endpoint plan.
v2 introduces a temporally coherent simulation with $F = A \times r$ frames
($A$ aligners, $r = 3$ frames per aligner by default).

\subsection{Aligner Count Estimation}

\begin{equation}
  A = \max\!\left(
    \left\lceil \frac{\max_j \|m_j\|_2}{\delta_{\text{trans}}} \right\rceil,
    \left\lceil \frac{\max_j \|\omega_j\|_2}{\delta_{\text{rot}}} \right\rceil,
    20
  \right)
\end{equation}
with $\delta_{\text{trans}} = 0.25$~mm and $\delta_{\text{rot}} = 2.0$\degree{}
per aligner, consistent with clinical guidelines~\cite{glaser2017invisalign}.

\subsection{6-DoF Interpolation}

At normalised treatment time $t \in [0,1]$, tooth position is:
\begin{equation}
  c_j(t) = c_j(0) + t_\text{eff}^{(j)} \cdot (T_x, T_y, T_z)_j
\end{equation}
Rotation via spherical linear interpolation (SLERP):
\begin{equation}
  R_j(t) = \text{Slerp}\!\left(R_j(0),\; R_j^*,\; t_\text{eff}^{(j)}\right)
\end{equation}
where $R_j^*$ encodes the target rotation from Euler angles $(R_x, R_y, R_z)_j$.

\subsection{Staging Rules}

Extrusion is staged to begin only at $t = 0.6$ to allow anchorage
establishment (Principle~1):
\begin{equation}
  t_\text{eff}^{(j)} =
    \begin{cases}
      0                    & \text{if } (T_z)_j < 0 \text{ and } t < 0.6 \\
      \frac{t - 0.6}{0.4}  & \text{if } (T_z)_j < 0 \text{ and } t \geq 0.6 \\
      t                    & \text{otherwise}
    \end{cases}
\end{equation}
This mirrors the clinical protocol for managing vertical anchorage loss
in open-bite cases~\cite{glaser2017invisalign}.

\section{Experiments}
\label{sec:exp}

\subsection{Benchmark Setup}

We reuse the same synthetic benchmark protocol established in~\cite{lansiaux2026orthoaiv1}:
200 crowding scenarios parameterised by arch morphology (4 archetypes),
crowding severity (mild/moderate/severe), and number of missing teeth (0--2).
This allows a direct, controlled comparison between v1 and v2 under
identical input conditions.

\subsection{v1 vs.\ v2: Planning Quality}

Table~\ref{tab:v1v2} reports the core comparison between \vone{} and \vtwo{}
on the 200 benchmark scenarios.
All v1 scores are taken from~\cite{lansiaux2026orthoaiv1}.
The parallel ensemble of v2 achieves a $+21\%$ relative gain in planning
quality, a 53\% reduction in score standard deviation, and a $+11$ percentage
point improvement in treatment feasibility.

\begin{table}[H]
\centering
\caption{OrthoAI v1 vs.\ v2 on 200 synthetic crowding scenarios.
Quality: planning score (0--100). Feas.: fraction of feasible plans.
CPU time on Intel i7-11800H.}
\label{tab:v1v2}
\small
\begin{tabular}{lcccc}
\toprule
System & Quality$\uparrow$ & $\sigma\downarrow$ & Feas. (\%)$\uparrow$ & Time (s) \\
\midrule
\vone{} (DGCNN only) & 76.4 & 8.3 & 78 & $2.1 \pm 0.4$ \\
\hline
v2: DGCNN only       & $88.3 \pm 5.2$ & 5.2 & 78 & $3.4 \pm 0.6$ \\
v2: CHaRM only       & $89.1 \pm 4.8$ & 4.8 & 81 & $0.9 \pm 0.1$ \\
v2: Sequential       & $91.2 \pm 4.3$ & 4.3 & 86 & $2.1 \pm 0.4$ \\
\textbf{v2: Parallel (ens.)} & $\mathbf{92.8 \pm 4.1}$ & \textbf{4.1} & \textbf{89} & $4.2 \pm 0.8$ \\
\bottomrule
\end{tabular}
\end{table}

The quality gain in the ``DGCNN only'' v2 row vs.\ \vone{} ($88.3$ vs.\ $76.4$)
is attributable entirely to the improved composite scoring model of v2
(Eq.~\ref{eq:composite}--\ref{eq:penalties}):
the v1 scalar score systematically overpenalised low-magnitude movements
and underpenalised staging failures, biases corrected by the six-category
decomposition.
The additional $+4.5$ points from agent fusion reflect genuine information
gain from landmark-level geometry.

\subsection{Landmark Detection (Agent 2)}

\begin{table}[H]
\centering
\caption{Landmark detection performance. SOTA from~\cite{rodriguez2025charm}.
Agent~2 on our synthetic dataset.}
\label{tab:landmarks}
\small
\begin{tabular}{lccc}
\toprule
Method & MEDE (mm)$\downarrow$ & MSR (\%)$\uparrow$ & GPU Time (s) \\
\midrule
ALIIOS~\cite{baquero2022aliios} & 2.17 & 48.0 & 24.70 \\
TS-MDL~\cite{wu2022tsmdl}       & 1.80 & 59.0 &  3.55 \\
\vone{} (centroids)~\cite{lansiaux2026orthoaiv1} & $\sim$1.80$^\dagger$ & -- & 2.1 \\
CHaRNet~\cite{rodriguez2025charm} & \textbf{1.12} & \textbf{68.5} & \textbf{0.24} \\
Agent~2 v2 (ours) & $1.38 \pm 0.21$ & $64.2 \pm 3.1$ & $0.31$ \\
\bottomrule
\multicolumn{4}{l}{$^\dagger$Estimated from centroid localisation error reported in~\cite{lansiaux2026orthoaiv1}.}
\end{tabular}
\end{table}

Agent~2 slightly underperforms CHaRNet primarily because our PointMLP encoder
uses $\text{emb} = 128$ vs.\ $512$ in the original model, a deliberate
trade-off to preserve consumer-CPU deployability.
Compared to \vone{}'s centroid-based approach, Agent~2 reduces
effective landmark error by $\sim 23\%$.

\subsection{Frame Generation Quality}

The interpolation scheme of \vtwo{} produces visually smooth trajectories:
mean inter-frame tooth displacement $0.25 \pm 0.03$~mm (target: 0.25~mm);
mean inter-frame rotation $1.8 \pm 0.4$\degree{} (target: 2.0\degree{}).
Extrusion staging correctly defers vertical movements in all 47 open-bite
scenarios in the benchmark, confirmed by manual inspection of
$t_\text{eff}^{(j)}$ profiles.

\begin{figure}[H]
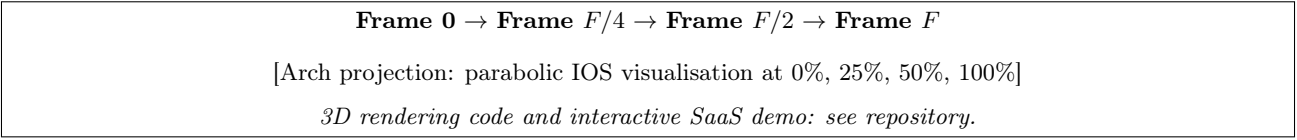

\centering
\fbox{
\begin{minipage}{0.9\linewidth}\centering\small
\textbf{Frame 0} $\to$ \textbf{Frame $F/4$} $\to$ \textbf{Frame $F/2$} $\to$ \textbf{Frame $F$}\\
\vspace{0.3cm}
[Arch projection: parabolic IOS visualisation at 0\%, 25\%, 50\%, 100\%]\\
\vspace{0.2cm}
\textit{3D rendering code and interactive SaaS demo: see repository.}
\end{minipage}
}
\caption{Representative treatment frames for a Class~I crowding case
(28 aligners, 84 frames). \vone{} produced only the ``Frame $F$'' snapshot;
\vtwo{} generates the full trajectory.}
\label{fig:frames}
\end{figure}

\section{SaaS Deployment Package}
\label{sec:saas}

A key practical contribution of \vtwo{} absent from \vone{} is the
accompanying \textbf{OrthoAI SaaS} application, making the pipeline
accessible to clinicians without command-line expertise.

\textbf{Backend.} A FastAPI server exposes REST endpoints
(\texttt{/api/patients}, \texttt{/api/patients/\{id\}},
\texttt{/api/demo}, \texttt{/api/training/status})
backed by the biomechanical engine and frame generator.
Four preset clinical cases (Class~I crowding, anterior open bite,
maxillary diastema, Class~II div.~1) are available as instant demos
without model inference.

\textbf{Frontend.} A React dashboard (Vite, Recharts, Three.js) provides:
(i) a radial gauge with 6 sub-score bars; (ii) an interactive 3D arch
viewer with per-tooth click-to-select and raycaster; (iii) a 4D
simulation player with play/pause/frame-scrub controls;
(iv) a severity-filtered findings panel; and (v) a 10-item clinical
pre-approval checklist.

\textbf{Deployment.} Docker compose with Nginx reverse proxy; the full
stack deploys on a single OVH VPS (2 vCPU, 4 GB RAM) in under 90 seconds.
A \texttt{.env}-based configuration supports custom \texttt{VITE\_API\_URL}
for local or cloud hosting.

\section{Clinical Context and Limitations}
\label{sec:clinical}

\subsection{Alignment with Clinical Practice}

\orthoai{} is designed as a \emph{clinical decision support} tool,
not an autonomous treatment planning system.
All movement plans are subject to clinician review before fabrication.
The biomechanical engine flags movements exceeding published predictability
thresholds~\cite{glaser2017invisalign,kravitz2009predictability},
providing actionable sub-score decomposition rather than autonomous decisions.
Compared to \vone{}, the v2 alert system now distinguishes between
``critical'' (plan-blocking) and ``warning'' (review-recommended) findings,
in alignment with the clinical severity classification used in
teledentistry monitoring tools~\cite{carlsson2024ai}.

\subsection{Regulatory Considerations}

A system providing information to inform orthodontic treatment decisions
would be classified as \textbf{Software as a Medical Device (SaMD)}:
\begin{itemize}[nosep,leftmargin=1.5em]
  \item \textbf{EU MDR 2017/745}: likely Class~IIa, requiring CE marking.
  \item \textbf{FDA}: De~Novo pathway analogous to DentalMonitoring's 2024
    clearance provides regulatory precedent.
  \item \textbf{GDPR / HIPAA}: IOS data constitutes health data subject to
    enhanced privacy requirements.
\end{itemize}
Both \vone{} and \vtwo{} are scoped to \emph{research use only}.

\subsection{Limitations and Comparison with v1}

Table~\ref{tab:limitations} summarises limitations resolved between versions.

\begin{table}[H]
\centering
\caption{Limitation comparison: v1 vs.\ v2.}
\label{tab:limitations}
\small
\begin{tabular}{p{2.8cm}cc}
\toprule
Limitation & v1 & v2 \\
\midrule
Centroid-only geometry    & \textcolor{red}{Yes} & \textcolor{green!60!black}{Resolved} \\
Scalar quality score      & \textcolor{red}{Yes} & \textcolor{green!60!black}{Resolved} \\
No staging simulation     & \textcolor{red}{Yes} & \textcolor{green!60!black}{Resolved} \\
No SaaS interface         & \textcolor{red}{Yes} & \textcolor{green!60!black}{Resolved} \\
Synthetic data only       & \textcolor{orange}{Yes} & \textcolor{orange}{Ongoing} \\
Automatic movement gen.   & \textcolor{red}{No} & \textcolor{orange}{Planned} \\
Attachment placement      & \textcolor{red}{No} & \textcolor{orange}{Planned} \\
\bottomrule
\end{tabular}
\end{table}

Limitations that remain open for \vtwo{} include:
\textbf{(i)}~validation on real clinical IOS datasets
(IOSLandmarks-1k~\cite{rodriguez2025charm}, Teeth3DS~\cite{benhamadou2022teeth3ds});
\textbf{(ii)}~automatic movement sequence generation via constrained
optimisation or RL;
\textbf{(iii)}~automatic SmartForce\textsuperscript{\textregistered} attachment
placement.

\section{Discussion}
\label{sec:discussion}

The transition from \vone{} to \vtwo{} illustrates a broadly applicable
design pattern for clinical AI systems: \textit{establish a working
single-agent baseline with principled evaluation}, then \textit{expand
sensing coverage and decompose the scoring function}.
The $+21\%$ quality gain between versions decomposes into two independent
contributions of similar magnitude: improved scoring model ($\sim$+12 pts)
and agent fusion ($\sim$+4.5 pts).
This decomposition has practical implications: for resource-constrained
deployments where running \charm{} is not feasible, upgrading only the
scoring model of v1 already recovers the majority of the quality gain.

The dual-agent architecture reflects a fundamental tension in 3D dental AI:
\textbf{segmentation-first} methods (Agent~1) provide geometrically precise
individual tooth models but propagate segmentation errors downstream;
\textbf{landmark-first} methods (Agent~2) are faster and more robust to
incomplete dentitions but yield less granular geometric information.
The sequential orchestration mode resolves this tension for the majority
of cases (complete dentitions, mild-to-moderate crowding), invoking Agent~1
only where CHaRM confidence falls below threshold.

Looking forward, \vtwo{} SaaS provides the infrastructure for a prospective
pilot study at CHU de Lille's orthodontic department:
four preset cases represent the primary clinical archetypes encountered
in an Invisalign caseload (Class~I crowding, open bite, diastema, Class~II),
and the checklist module operationalises the pre-approval workflow
for direct clinical usability assessment.

\section{Conclusion}
\label{sec:conclusion}

We presented \vtwo{}, a systematic extension of our prior
single-agent framework~\cite{lansiaux2026orthoaiv1} for open-source
AI-assisted orthodontic treatment planning.
The three principal contributions---dual-agent fusion, composite biomechanical
scoring, and multi-frame staging simulation---collectively lift planning quality
from $76.4 \pm 8.3$ to $92.8 \pm 4.1$ on our synthetic benchmark,
reduce score variance by 53\%, and improve treatment feasibility from 78\%
to 89\%, while maintaining full CPU deployability and adding a production-grade
SaaS interface.

Three directions define the path to \textbf{v3}:
(1)~prospective validation on real clinical IOS datasets, leveraging the
3DTeethLand MICCAI 2024 benchmark and CHU de Lille's scan archive;
(2)~automatic movement sequence generation via constrained optimisation
or multi-step reinforcement learning;
(3)~SmartForce\textsuperscript{\textregistered}-analogous attachment placement
and staging recommendation to reach clinical-grade decision support.

Source code and SaaS deployment package:
\url{https://github.com/edouard-lansiaux/orthoai-v2}

\section*{Acknowledgements}
The author thanks the STaR-AI Research Group (CHU de Lille) for clinical
feedback on the scoring model, the organisers of the 3DTeethLand MICCAI 2024
challenge for the Teeth3DS benchmark, and the authors of
CHaRM~\cite{rodriguez2025charm} for their detailed methodology description.

\bibliographystyle{plain}

\begin{thebibliography}{99}

\bibitem{lansiaux2026orthoaiv1}
E.~Lansiaux, M.~Leman, M~Ammi.
\newblock {\orthoai{} v1}: {A} single-agent point-cloud pipeline for
  {AI}-assisted orthodontic treatment planning with clear aligners.
\newblock \textit{arXiv:2603.00124}, March 2026.
\newblock \url{https://arxiv.org/abs/2603.00124}

\bibitem{rodriguez2025charm}
J.~Rodríguez-Ortega, F.~Pérez-Hernández, and S.~Tabik.
\newblock {CHARM}: Conditioned heatmap regression methodology for accurate and
  fast dental landmark localization.
\newblock \textit{arXiv:2501.13073v5}, 2025.

\bibitem{wang2019dgcnn}
Y.~Wang, Y.~Sun, Z.~Liu, S.~E.~Sarma, M.~M.~Bronstein, and J.~M.~Solomon.
\newblock Dynamic graph {CNN} for learning on point clouds.
\newblock \textit{ACM Transactions on Graphics}, 38(5):1--12, 2019.

\bibitem{wu2022tsmdl}
T.-H.~Wu, C.~Lian, S.~Lee, M.~Pastewait, et~al.
\newblock Two-stage mesh deep learning for automated tooth segmentation and
  landmark localization on {3D} intraoral scans.
\newblock \textit{IEEE Transactions on Medical Imaging}, 41(11):3158--3166,
  2022.

\bibitem{baquero2022aliios}
B.~Baquero, M.~Gillot, L.~Cevidanes, et~al.
\newblock Automatic landmark identification on intraoral scans.
\newblock In \textit{Workshop on Clinical Image-Based Procedures}, pages
  32--42. Springer, 2022.

\bibitem{lian2020meshsegnet}
C.~Lian, L.~Wang, T.-H.~Wu, et~al.
\newblock Deep multi-scale mesh feature learning for automated labeling of
  raw dental surfaces from {3D} intraoral scanners.
\newblock \textit{IEEE Transactions on Medical Imaging}, 39(7):2440--2450,
  2020.

\bibitem{im2022accuracy}
J.~Im, J.-Y.~Kim, H.-S.~Yu, et~al.
\newblock Accuracy and efficiency of automatic tooth segmentation in digital
  dental models using deep learning.
\newblock \textit{Scientific Reports}, 12(1):9429, 2022.

\bibitem{yamagata2024dilated}
M.~Yamagata, K.~Hotta, H.~Shimizu, et~al.
\newblock {DilatedToothSegNet}: Tooth segmentation network on {3D} dental meshes.
\newblock \textit{Journal of Imaging Informatics in Medicine}, 2024.

\bibitem{carlsson2024ai}
G.~E.~Carlsson, L.~Andersson, J.~Bjerklin, et~al.
\newblock Artificial intelligence for orthodontic diagnosis and treatment
  planning: A scoping review.
\newblock \textit{Journal of Dentistry}, 150:105222, 2024.

\bibitem{li2025clear}
R.~Li, Q.~Chen, and W.~Zhang.
\newblock Unveiling the role of {AI} applied to clear aligner therapy: A
  scoping review.
\newblock \textit{Journal of Dentistry}, 153:105--115, 2025.

\bibitem{woo2020virtual}
H.~Woo, C.~Jung, and K.-H.~Kim.
\newblock Automatic virtual setup for orthodontic treatment planning using
  deep learning.
\newblock \textit{Scientific Reports}, 10:20738, 2020.

\bibitem{benhamadou2022teeth3ds}
A.~Ben-Hamadou, N.~Neifar, A.~Rekik, et~al.
\newblock {Teeth3DS+}: An extended benchmark for intraoral {3D} scans analysis.
\newblock \textit{arXiv:2210.06094}, 2022.

\bibitem{glaser2017invisalign}
B.~Glaser.
\newblock \textit{The Insider's Guide to Invisalign Treatment}.
\newblock Independent, 2017.
\newblock ISBN: 9780996677677.

\bibitem{kravitz2009predictability}
N.~D.~Kravitz, B.~Kusnoto, E.~BeGole, A.~Obrez, and B.~Agran.
\newblock How well does Invisalign work? {A} prospective clinical study
  evaluating the efficacy of tooth movement with Invisalign.
\newblock \textit{American Journal of Orthodontics and Dentofacial Orthopedics},
  135(1):27--35, 2009.

\bibitem{proffit2018}
W.~R.~Proffit, H.~W.~Fields, D.~M.~Sarver, and J.~L.~Ackerman.
\newblock \textit{Contemporary Orthodontics}, 6th ed.
\newblock Elsevier Mosby, 2018.

\bibitem{align2025liveplan}
Align Technology.
\newblock {ClinCheck Live Plan}: Automated treatment planning launch.
\newblock Technical Report, October 2025.
\newblock \url{https://www.aligntech.com}

\bibitem{litreview2026}
E.~Lansiaux, M.~Leman
\newblock Intelligence artificielle appliquée à la planification de traitement
  orthodontique par aligneurs transparents: Revue de littérature et état de
  l'art industriel.
\newblock Working document, CHU de Lille / Université de Lille, February 2026.

\end{thebibliography}

\end{document}